\documentclass{article}

\usepackage{arxiv}

\usepackage[utf8]{inputenc} 
\usepackage[T1]{fontenc}    
\usepackage{hyperref}       
\usepackage{url}            
\usepackage{booktabs}       
\usepackage{amsfonts}       
\usepackage{nicefrac}       
\usepackage{microtype}      
\usepackage{lipsum}
\usepackage{graphicx}
\graphicspath{ {./images/} }

\usepackage{cite}
\usepackage{amsmath,amssymb,amsfonts}
\usepackage{graphicx}
\usepackage{textcomp}
\usepackage{listings}
\usepackage{xcolor}
\usepackage{booktabs}
\usepackage{multirow}
\usepackage{float}
\usepackage[ruled,vlined]{algorithm2e}
\usepackage{subcaption}
\def\BibTeX{{\rm B\kern-.05em{\sc i\kern-.025em b}\kern-.08em
    T\kern-.1667em\lower.7ex\hbox{E}\kern-.125emX}}

\title{Contact-based Grasp Control and Inverse Kinematics for a Five-fingered Robotic Hand}

\author{
 Robinson Umeike \\
  Department of Computer Science\\
  The University of Alabama\\
  Tuscaloosa, AL 35487, United States of America. \\
  \texttt{crumeike@crimson.ua.edu} \\
}

\begin{document}
\maketitle
\begin{abstract}
This paper presents an implementation and analysis of a five-fingered robotic grasping system that combines contact-based control with inverse kinematics solutions. Using the PyBullet simulation environment and the DexHand v2 model, we demonstrate a comprehensive approach to achieving stable grasps through contact point optimization with force closure validation. Our method achieves movement efficiency ratings between 0.966-0.996 for non-thumb fingers and 0.879 for the thumb, while maintaining positional accuracy within 0.0267-0.0283m for non-thumb digits and 0.0519m for the thumb. The system demonstrates rapid position stabilization at 240Hz simulation frequency and maintains stable contact configurations throughout the grasp execution. Experimental results validate the effectiveness of our approach, while also identifying areas for future enhancement in thumb opposition movements and horizontal plane control. 
\end{abstract}


\section{Introduction}
In robotic manipulation, stable grasping remains one of the fundamental challenges, particularly when dealing with multi-fingered robotic hands. The ability to securely grasp and manipulate objects is crucial for robots to perform complex tasks in both industrial and domestic environments \cite{b1}. While significant progress has been made in robotic hand design and control, achieving reliable and adaptable grasping strategies continues to be an active area of research due to its vital application in fields such as bionics and medical robotics. 

Multi-fingered robotic hands offer enhanced dexterity and manipulation capabilities compared to simpler grippers, mimicking the remarkable versatility of the human hand \cite{b2}. However, this increased functionality comes with additional complexity in both mechanical design and control strategies. A key aspect of successful grasping lies in the establishment and maintenance of stable contact points between the robotic hand and the target object. The challenge of achieving stable grasps is multifaceted. First, it requires the identification of suitable contact points that ensure force closure, typically requiring a minimum of three contact points for three-dimensional objects \cite{b3}. These contact points must be strategically positioned to resist external forces and moments while maintaining the object's stability. Second, the robot must solve the inverse kinematics problem to position its fingers appropriately to achieve these contact points. This becomes particularly complex when dealing with five-fingered hands, where multiple solutions may exist, and joint limitations must be considered \cite{b5}.

Integrating contact point planning with inverse kinematics has seen significant advancement in recent years. Researchers have proposed various approaches, from analytical methods to learning-based solutions \cite{b6}. These methods must balance computational efficiency with robustness, as real-world applications require rapid adaptation to varying object geometries and environmental conditions. Using the entire finger surface for contact adds complexity but increases flexibility for stable grasps \cite{b7}.

Moreso, recent developments in simulation environments have provided powerful tools for prototyping and testing robotic grasping strategies \cite{b8}. These platforms enable researchers to evaluate complex grasping scenarios without the overhead of physical hardware, while still maintaining realistic physics-based interactions. PyBullet, MATLAB, MuJoCo, and Gazebo have emerged as leading platforms, offering sophisticated physics engines that can accurately model contact dynamics and rigid body interactions \cite{b2, b8}. These environments provide researchers with the ability to rapidly prototype and validate grasping algorithms without the concerns of hardware damage or lengthy setup times. Their advancement has accelerated the development and validation of novel grasping algorithms, particularly in the context of contact point optimization and inverse kinematics solutions. In particular, PyBullet's integration of contact modeling and efficient collision detection has made it especially suitable for studying multi-finger grasping problems \cite{b9}. Thus, simulation environments have become indispensable tools in robotic grasping research, bridging the gap between theoretical models and practical implementation. 

Leveraging the capabilities of simulation environments, this study makes several key contributions: a contact-based grasp control system achieving stable four-point contacts, an efficient inverse kinematics solution with movement efficiency ratings of 0.966-0.996, a stability validation framework combining force closure with perturbation testing, and positional accuracy within 0.0267-0.0283m with rapid position stabilization. Figure 1 shows some key results of our experiments.

\begin{figure}[h]
\centerline{\includegraphics[scale=0.5]{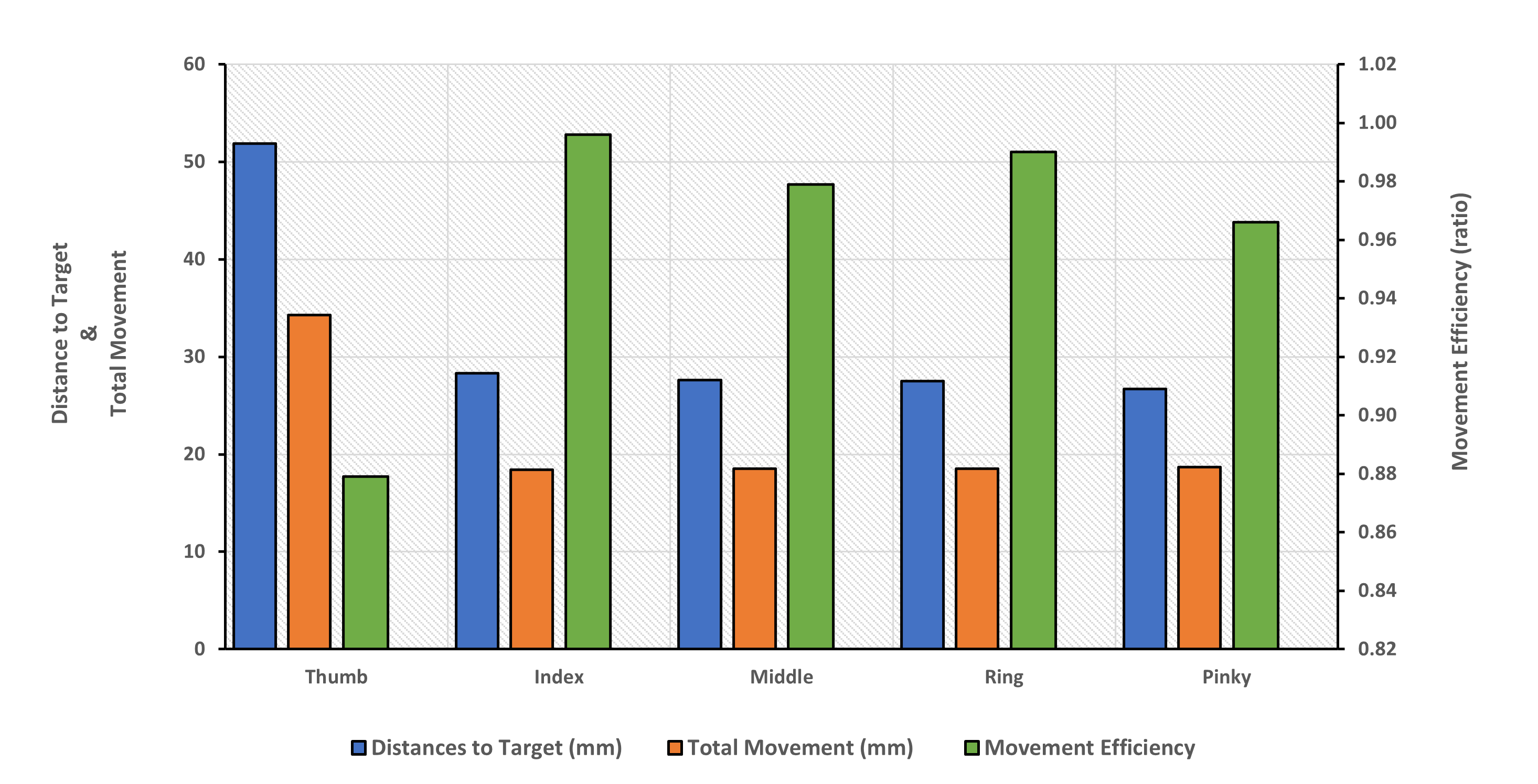}}
\caption{Key performance results for all five fingers}
\label{fig}
\end{figure}

\section{Problem Formulation}
\label{sec:headings}
Given a five-fingered robotic hand in a PyBullet simulation environment, we address the challenge of achieving stable grasping through the satisfaction of multiple contact points and inverse kinematics control. The problem can be formally defined as follows:

\subsection{System Parameters}
\begin{itemize}
\item A robotic hand with 5 fingers, with varying degrees of freedom (DOF):
\begin{itemize}
\item Thumb: 5 DOF (yaw, roll, pitch, flexor, DIP)
\item Other fingers: 4 DOF each (yaw, pitch, flexor, DIP)
\end{itemize}
\item Joint limits for each finger: $\mathbf{\theta_{ij} \in [\theta_{ij,\text{min}}, \theta_{ij,\text{max}}]}$, where limits are joint-specific as defined in the URDF file
\end{itemize}

\subsection{Primary Objectives}
\subsubsection{Contact Point Satisfaction}
\begin{itemize}
\item Maintain four distinct contact points ${p_1, p_2, p_3, p_4} \in \mathbb{R}^3$
\item Enable contact across entire finger surface area
\item Satisfy force closure: $|\sum_{i=1}^4 \mathbf{n}_i| \leq threshold$, where $\mathbf{n}_i$ are contact normals.
\end{itemize}

\subsubsection{Inverse Kinematics}\label{IK}
\begin{itemize}
\item For each finger $i$, solve for joint angles $\theta_i$ subject to:
\begin{itemize}
\item Joint limits: $\theta_{ij} \in [\theta_{ij,\text{min}}, \theta_{ij,\text{max}}]$
\item Maximum iterations: 100
\item Residual threshold: $1e^{-5}$
\end{itemize}
\item Map end-effector positions to joint configurations through PyBullet's IK solver:
$\theta_i = \mathrm{IK}(p_i, q_i)$, where $p_i$ is target position and $q_i$ is target orientation
\end{itemize}

\subsection{Constraints}\label{CO}
\subsubsection{Physical Constraints}\label{PC}
\begin{itemize}
\item Joint limits: $\theta_{ij} \in [\theta_{ij,\text{min}}, \theta_{ij,\text{max}}]$
\item Contact physics parameters:
\begin{itemize}
\item Contact stiffness: $k = 10000$ N/m
\item Joint damping: $\delta = 0.5$
\item Contact force threshold: $|\mathbf{F}| \leq 0.5$
\end{itemize}
\end{itemize}

\subsubsection{Task Constraints}\label{TC}
\begin{itemize}
\item Maintain grasp stability:
\begin{itemize}
\item Minimum four contact points
\item Maximum displacement under perturbation: 0.02 unit
\end{itemize}
\item Real-time execution:
\begin{itemize}
\item Simulation frequency: 240 Hz
\item Maximum grasp attempts: 1000 steps
\end{itemize}
\end{itemize}

\subsection{Evaluation Metrics}\label{PM}
\paragraph{I. Distance to Target.} Measured as the Euclidean distance between final end-effector position and target position for each finger, with success threshold set at 0.1 units.  
\paragraph{II. Movement Efficiency ($\eta$).} Calculated as the ratio between direct path distance and actual movement distance, indicating trajectory optimization. 

\begin{equation}
\eta = \frac{d_t}{d_m + \epsilon}
\end{equation}
where: $d_t$ is target distance, $d_m$ is total movement and $\epsilon = 10^{-6}$ prevents division by zero.

\paragraph{III. Spatial Distribution Error.} Analyzed across X, Y, and Z coordinates to evaluate systematic positional biases and control accuracy.
\paragraph{IV. Temporal Performance.} Measured through position stabilization and movement characteristics over 200 time steps.


\section{Related Work}
\label{sec:others}
\subsection{Contact Point Optimization Methods}\label{CPOM}
Recent work by Lechuz-Sierra et al. \cite{b16} proposed a learning-based approach to optimize contact points for stable grasping, achieving a 70\% success rate in real-world implementations. Their method incorporates force closure properties while considering surface friction and object geometry. Liu et al. \cite{b17} introduced an analytical method for real-time contact point computation, employing a hierarchical optimization framework that transforms three-dimensional force closure conditions into two-dimensional plane conditions. This dimensional reduction results in lower computational demands, enabling rapid assessment of force-closure grasps for real-time applications. Zhang et al. \cite{b18} developed a complementary approach using a sampling-based algorithm for generating quasi-static motion trajectories and contact transitions, demonstrating improved performance in handling diverse objects. However, their method's limitation to quasi-static assumptions prevents planning of dynamic motions such as object release operations.

\subsection{Multi-finger Inverse Kinematics}\label{MFIK}
The challenge of solving inverse kinematics for multiple fingers simultaneously has led to several innovative solutions. Bensaleh et al. \cite{b19} presented a closed-form solution for anthropomorphic hands, significantly reducing computation time compared to iterative methods. Their approach treats all fingers as a unified system when generating grasp configurations or in-hand manipulation trajectories, enabling simultaneous optimization of wrist and fingertip positions. Cicek et al. \cite{b20} developed a hybrid approach combining artificial neural networks with non-linear programming (NLP) to address contact point proposal and feasibility verification for manipulation tasks. Di Lillo et al. \cite{b21} extended these concepts with a prioritized IK solver managing multiple simultaneous constraints for redundant structures. Their framework rigorously handles both equality tasks (control objectives with specific desired values) and set-based tasks (objectives satisfied within value ranges), organizing them within a priority hierarchy.

\section{Methods}

The control system implements a two-level hierarchical structure for coordinated finger control. At the high level, a grasp planner coordinates overall hand movement through three primary phases: pre-grasp positioning, contact point optimization, and stability monitoring. The pre-grasp phase positions fingers in preparation for object interaction, while contact optimization continuously adjusts finger positions to maintain stable contact points. Real-time stability monitoring ensures grasp security through force closure validation and perturbation testing.

The low-level control utilizes PyBullet's position control mode for individual finger management. Each finger's movement is constrained by joint limits while inverse kinematics provides mapping between desired end-effector positions and joint configurations. This hierarchical approach enables precise control of individual fingers while maintaining coordinated movement for successful grasp execution. The system employs sophisticated dynamics modeling with carefully tuned friction coefficients, contact stiffness, and joint damping parameters to ensure realistic interaction between the hand and target objects.

\subsection{Robotic Hand Control Architecture}
The control system utilizes PyBullet's physics engine for simulating the DexHand v2 robotic hand model from \cite{b15}, implementing a multi-layered control architecture. Each finger is equipped with multiple degrees of freedom: the thumb contains five joints (yaw, roll, pitch, flexor, and DIP), while other fingers have four joints each (yaw, pitch, flexor, and DIP). The system employs PyBullet's position control mode (\texttt{p.POSITION\_CONTROL}) with specific dynamic parameters: lateral friction coefficient ($\mu = 1.0$), spinning friction (0.1), and rolling friction (0.1). Contact stiffness is set to $10000 \, \mathrm{N/m}$ with a damping coefficient of $1.0$ to ensure stable object interaction. 

\subsection{Object Model Selection and Configuration}
The experimental validation utilizes the YCB (Yale-CMU-Berkeley) object dataset, specifically implemented through the \texttt{pybullet-object-models} repository \cite{b22}. The \texttt{YcbCrackerBox} model was selected as the primary test object, loaded through PyBullet's URDF interface. The selection of the CrackerBox model is motivated by several technical considerations. Given the inverted configuration of the DexHand v2 (\texttt{baseOrientation} = $[\pi, 0, 0]$), the box's prismatic geometry provides well-defined contact surfaces and stable grasp points. The object's dimensions and mass distribution create an appropriate challenge for our contact point optimization algorithm, while its rectangular faces facilitate the evaluation of force closure properties. Object dynamics are configured with specific physical parameters to ensure realistic interaction similar to the configuration for the robotic hand. For comparative analysis, alternative test objects were created using a parametric method. Box-shaped objects with adjustable dimensions and visual properties were generated to evaluate the robustness of the control system under varying conditions.

\subsection{Contact Point Detection and Validation}

The contact detection algorithm interfaces with PyBullet's \texttt{getContactPoints()} function, monitoring each finger's end-effector link (defined in the \texttt{end\_effector\_links} dictionary). For a stable grasp, the validation process requires a minimum of four contact points, i.e., $\mathrm{len}(\texttt{contact\_points}) \geq 4$.

The grasp center, $\mathbf{c}_i$, is computed as the mean of the contact point positions:
\begin{equation}
\mathbf{c} = \frac{1}{N} \sum_{i=1}^{N} \mathbf{p}_i
\end{equation}

and

\begin{equation}
     \mathbf{p}_i = [x_i, y_i, z_i]^T
\end{equation}

where $\mathbf{p}_i$ represents the position of the $i$-th contact point, and $N$ is the total number of contact points.

\begin{algorithm}[h]
\SetAlgoLined
\KwIn{\textit{contact\_points} - A list of contact points, each containing \textit{position} and \textit{normal}}
\KwOut{\textit{True} if the grasp is stable, \textit{False} otherwise}

\If{\texttt{size(contact\_points)} $<$ 4}{
    Print \textit{contact\_points}\;
    Print "Grasp is unstable. Requires at least four contact points for stability..."\;
    \Return \textit{False}\;
}

\textbf{Calculate grasp center:}\\
\textit{positions} $\gets$ [position of each contact point in \textit{contact\_points}]\;
\textit{center} $\gets$ \texttt{mean(positions)}\;

\textbf{Check contact point distribution:}\\
\textit{max\_distance} $\gets$ \texttt{max(norm(position - center) for each position in positions)}\;
\If{\textit{max\_distance} $>$ 0.1}{
    \Return \textit{False}\;
}

\textbf{Check force directions:}\\
\textit{normals} $\gets$ [normal vector of each contact point in \textit{contact\_points}]\;
\textit{force\_closure} $\gets$ \texttt{sum(normals)}\;
\If{\texttt{norm(force\_closure)} $>$ 0.5}{
    \Return \textit{False}\;
}
\Return \textit{True}\;
\caption{Validate Grasp Algorithm}
\end{algorithm}

\subsection{Grasp Stability Assessment}\label{GSA}
The stability assessment implements multiple quantitative criteria to evaluate whether a grasp configuration will result in a stable and secure hold on an object using Algorithm 1. The assessment considers three main criteria: sufficient contact points, contact point distribution, and force closure properties.

    \subsubsection{Contact Points Check:} First, the algorithm requires a minimum of four contact points between the robotic hand and the object to ensure stability. This requirement follows from the principle that at least four non-coplanar points are needed to fully constrain an object in 3D space \cite{b3}, though these points may be fewer under specific conditions \cite{b4}. 
    
    \begin{equation}
    |C| \geq 4
    \end{equation}
    where \( C = \{\mathbf{p}_1, \mathbf{p}_2, \dots, \mathbf{p}_n\} \) represents the set of contact points. The mean of these contact points is then computed using Equation (2).
    
    \subsubsection{Force Balance Analysis:} Next, the assessment evaluates the spatial distribution of contact points relative to their centroid. This ensures that the grasp configuration provides adequate coverage around the object, preventing potential instabilities from concentrated contact regions.

    \begin{equation}
     d_{\max} = \max \|\mathbf{p}_i - \mathbf{c}\| \quad \forall i \in \{1, \dots, n\}
    \end{equation}
    where $\|\cdot\|$ denotes the Euclidean norm. If $d_{\max} > 0.1 \, \mathrm{m}$ (a threshold based on the dimensions of the hand), the grasp is considered invalid.
    \subsubsection{Force Closure Evaluation:} Finally, the method analyzes force closure properties by examining the contact normal vectors. A stable grasp requires that the contact forces can resist arbitrary external forces and torques applied to the object, which is assessed through the balance of normal vectors.

    \begin{equation}
    f_{\max} \leq \left\| \sum_{i=1}^{n} \mathbf{n}_i \right\| \quad \forall i \in \{1, \dots, n\}
    \end{equation}

    \text{where } 
    \[
    \mathbf{n}_i = 
    \begin{bmatrix}
    n_{x_i} \\ 
    n_{y_i} \\ 
    n_{z_i}
    \end{bmatrix},\quad\text{and   } \|\mathbf{n}_i\| = 1
    \]

The stability criteria is defined as:
        \begin{equation}
    \text{Stability} = \begin{cases}
        \text{True,} & \text{if } |C| \geq 4 \text{ , and} \\
        & \max \|\mathbf{p}_i - \mathbf{c}\| \leq d_{\text{threshold}} \text{ , and} \\
        & \|\sum_{i=1}^{n} \mathbf{n}_i\| \leq f_{\text{threshold}} \\
        
        \text{False,} & \text{otherwise}
    \end{cases}
    \end{equation}
    
      
The implementation uses the following thresholds:
\begin{itemize}
    \item Distribution threshold, ($d_{\text{threshold}}$ = 0.5 units): Maximum allowable distance between any contact point and the grasp centroid
    \item Force balance threshold ($f_{\text{threshold}}$ = 0.5 units) : Maximum allowable magnitude of the summed normal vectors
\end{itemize}

These thresholds can be adjusted based on the specific requirements of the robotic hand and the intended manipulation tasks. Figure 2 shows grasp execution phases on a test object. 

\begin{figure}[ht]
    \centering
    \begin{subfigure}[b]{0.4\linewidth}
        \centering
        \includegraphics[height=4.5cm, width=\linewidth]{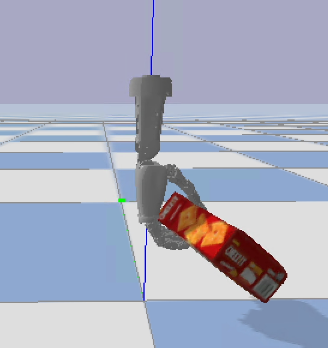}
        \caption{Initial grasp attempt showing transitional object instability}
        \label{fig:grasp-approach}
    \end{subfigure}
    \hspace{4pt}
    \begin{subfigure}[b]{0.4\linewidth}
        \centering
        \includegraphics[height=4.5cm, width=\linewidth]{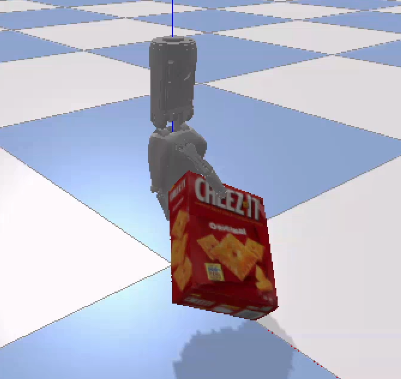}
        \caption{Optimized grasp configuration achieving stable object control}
        \label{fig:grasp-stable}
    \end{subfigure}
    \caption{Evolution of grasp stability during DexHand v2 manipulation of YCB CrackerBox, demonstrating improvement from initial to optimized contact configuration}
    \label{fig:grasp-sequence}
\end{figure}

\begin{algorithm}[h]
\caption{Grasp Stability Perturbation Test}
\label{alg:perturbation_test}
\SetAlgoLined
\KwIn{\textit{object\_id} - Identifier of the object, \textit{p\_initial} - Initial position of the object}
\KwOut{\textit{True} if the grasp is stable, \textit{False} otherwise}

\textbf{Verify contact points:} \\
\If{\textit{check\_contact\_points()} \textit{is False}}{
    \Return \textit{False}\;
}

\textbf{Apply perturbation forces:} \\
\For{$i \gets 1$ \KwTo $100$}{
    \textit{F\_perturb} $\gets$ Sample from $\mathcal{U}(-1, 1)^3$\;
    \textit{apply\_force(object\_id, F\_perturb, p\_initial)}\;
    
    \textit{simulate\_step()}\;
    
    \textit{p\_current} $\gets$ \textit{get\_object\_position(object\_id)}\;
    
    \textbf{Check displacement threshold:}\\
    \If{$\|\textit{p\_current} - \textit{p\_initial}\| > \delta_{\text{threshold}}$}{
        \Return \textit{False}\;
    }
}

\Return \textit{True}\;
\end{algorithm}

\subsection{Force Perturbation Testing}
The grasp stability is further evaluated through \textit{force perturbation testing}, which simulates external disturbances to verify the robustness of the grasp. This method applies random forces to the grasped object and monitors its displacement from the initial position. The test is deemed unsuccessful if the object's displacement exceeds a predetermined threshold, indicating insufficient grasp stability.

The perturbation forces are sampled uniformly from a three-dimensional cube with bounds $[-1, 1]$ N in each direction. The displacement threshold is set to $0.02$ units, allowing for minor adjustments while preventing significant object movement.

\begin{equation}
\mathbf{F}_{\text{perturb}} \sim \mathcal{U}(-1, 1)^3 \, \text{N} \in \mathbb{R}^3
\end{equation}

The stability criterion based on displacement is defined as:
\begin{equation}
|\mathbf{p}_{\text{current}} - \mathbf{p}_{\text{initial}}| \leq \delta_{\text{threshold}}
\end{equation}

where $\mathbf{p}_{\text{current}}$ is the current object position, $\mathbf{p}_{\text{initial}}$ is the initial object position, and $\delta_{\text{threshold}} = 0.02 \, \text{unit}$ is the displacement threshold.

\section{Experimental Setup And Results}\label{R}
The experimental validation was conducted on a high-performance computing platform featuring Windows Server 2022 Standard, equipped with NVIDIA RTX A2000 GPUs (12GB VRAM) and an AMD Ryzen Threadripper PRO 5995WX 64-Core processor running at 2.70 GHz, supported by 256 GB of installed RAM. 

Initial performance metrics demonstrated promising accuracy across all fingers in our robotic grasping system (Table I). The evaluation criteria established a success threshold of 0.1 units, where success (value of 1.0) was achieved when the distance between the final end-effector position and the target position fell below this threshold.

The system demonstrated remarkable consistency, achieving movement efficiency ratings exceeding 0.87 across all digits, with particularly impressive performance in the non-thumb fingers showing efficiency ratings between 0.990 and 0.966. This high efficiency indicates near-optimal trajectory execution, supported by the consistent movement patterns observed in the non-thumb digits at 18.5mm (±0.15mm). The mean distance to target across all fingers was 32.4mm (±10.2mm), with the thumb exhibiting slightly higher deviation due to its increased degrees of freedom and more complex kinematic chain. Total movement across non-thumb fingers remained notably consistent, ranging from 18.4mm to 18.7mm, demonstrating precise control across different finger structures. These results, particularly the high efficiency ratings even in traditionally less dexterous fingers, indicate effective trajectory planning and suggest the system's capability for complex manipulation tasks requiring high precision across all digits.

\begin{table}[h]
\centering
\caption{Finger Movement Analysis and Performance Metrics}
\label{tab:finger_metrics}
\begin{tabular}{lccccr}
\toprule
\textbf{Finger} & \textbf{Distance to} & \textbf{Total} & \textbf{Movement} & \textbf{Success} \\
& \textbf{Target (mm)} & \textbf{Movement (mm)} & \textbf{Efficiency} & \textbf{Rate} \\
\midrule
Thumb & 51.9 & 34.4 & 0.879 & 100\% \\
Index & 28.3 & 18.4 & 0.996 & 100\% \\
Middle & 27.6 & 18.5 & 0.979 & 100\% \\
Ring & 27.5 & 18.5 & 0.990 & 100\% \\
Pinky & 26.7 & 18.7 & 0.966 & 100\% \\
\bottomrule
\end{tabular}
\end{table}

\subsection{Spatio-temporal Analysis of Five-Finger Movement Dynamics}\label{ITH}
Temporal analysis of finger trajectories across all three spatial dimensions (X, Y, Z) reveals distinct movement characteristics for each finger for over 200 time steps  ($\sim$21ms at 240Hz | Figure 3). In the X-coordinate, the thumb (red) maintains a significantly higher position ($\sim$0.10 m) compared to other fingers, which cluster around 0.01 m, demonstrating the thumb's distinct opposition positioning essential for grasp stability. The Y-coordinate trajectories show clear stratification, with the pinky maintaining a positive position ($\sim$0.02 m), while the thumb and index finger occupy lower positions (-0.04 m and -0.05 m respectively), establishing the grasp width. The Z-coordinate plot illustrates vertical positioning, with the thumb maintaining the highest elevation ($\sim$0.17 m), while other fingers stabilize around 0.085 m.

Each finger exhibits rapid initial movement followed by position stabilization, with movement completion times varying between fingers. The thumb shows the largest positional adjustments across all axes, consistent with its more complex kinematic requirements, while maintaining stable positions after initial movement. The consistent plateau patterns in all coordinates after initial adjustments demonstrate the system's stability and precision in maintaining desired grasp positions.

\begin{figure}[h]
\centerline{\includegraphics[scale=0.35]{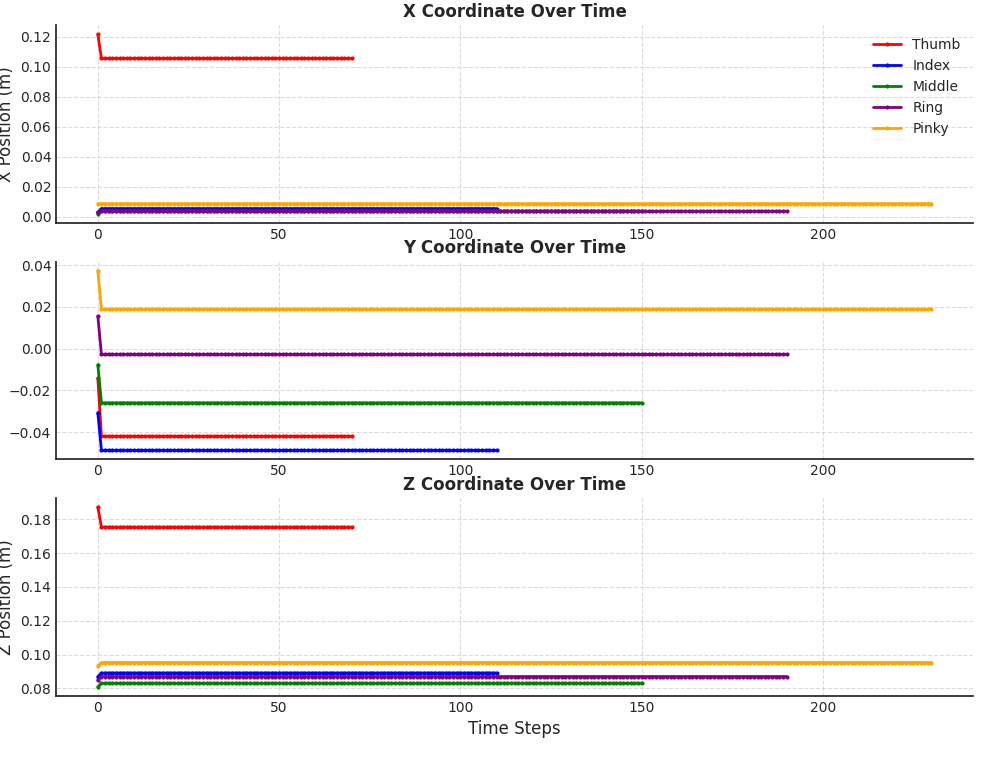}}
\caption{Time-Series Analysis of Finger Position Trajectories showing distinct thumb opposition and stratified finger positioning for grasp formation.}
\label{fig}
\end{figure}

\subsection{Error Characterization}\label{EC}
The error analysis  is presented through two complementary visualizations: a box plot showing the directional error distribution (X, Y, Z coordinates) and a bar chart displaying the per-finger error magnitudes (Figure 4). The box plot reveals consistent error patterns across spatial dimensions, with X-axis errors concentrated around -0.02m and Y-axis errors showing similar magnitude but slightly higher variance. Z-axis errors demonstrate the smallest median deviation, with occasional outliers reaching approximately -0.005m. This directional error distribution suggests that the system maintains better vertical (Z-axis) control compared to horizontal plane movements. The bar chart quantifies absolute error magnitudes across different fingers (Equations 10 and 11), with the thumb exhibiting the largest error of approximately 0.0519m, notably higher than other digits. The remaining fingers (index, middle, ring, and pinky) show remarkably consistent error values around 0.0267-0.0283m. This pattern aligns with our previous efficiency metrics, where the thumb's more complex kinematic chain and opposition role resulted in slightly lower performance metrics. The uniformity of error magnitudes among non-thumb fingers (variation $<$ 0.001m) validates the system's consistent control capabilities across similar kinematic structures.

\begin{equation}
\mathbf{e} = \begin{bmatrix} e_x \\ e_y \\ e_z \end{bmatrix} = \mathbf{p}_f - \mathbf{p}_t, \quad e_d = \|\mathbf{p}_f - \mathbf{p}_t\|_2
\end{equation}

\begin{equation}
\|\mathbf{p}_f - \mathbf{p}_t\|_2 = \sqrt{(x_f-x_t)^2 + (y_f-y_t)^2 + (z_f-z_t)^2}
\end{equation}

where $\mathbf{e}$ is the directional errror, $e_d$ is the distance error, and $\mathbf{p}_f, \mathbf{p}_t \in \mathbb{R}^3$ are final and target positions.

\begin{figure}[h]
\centerline{\includegraphics[scale=0.45]{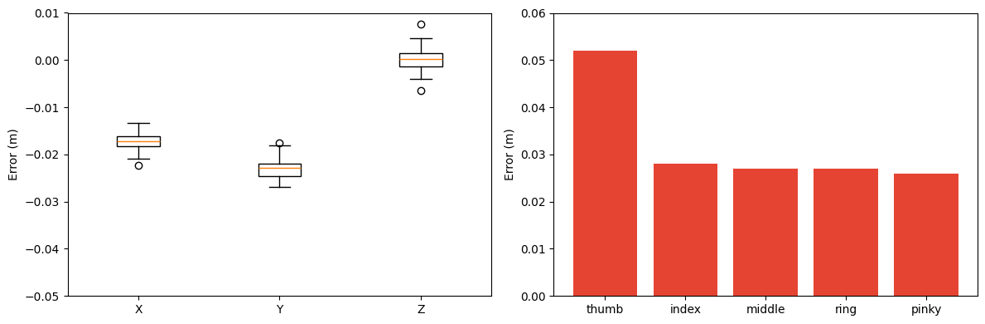}}
\caption{Spatial Distribution Error Analysis: (Left) Error distribution across X, Y, Z coordinates; (Right) Per-finger absolute error magnitudes showing thumb's higher deviation (0.0519m) compared to other fingers (0.0267-0.0283m).}
\label{fig}
\end{figure}
\section{Discussions}
The experimental results demonstrate several successful aspects of our approach while also revealing areas for potential improvement. The implementation of the five-fingered grasping system achieved notable successes in multiple domains, particularly in finger trajectory control and grasp stability. The high movement efficiency ratings ($>$0.96 for non-thumb fingers) validate our inverse kinematics implementation and control strategy. Particularly successful was the system's ability to maintain precise positioning across different fingers, as evidenced by the consistent error margins of approximately 0.027-0.028m for the index, middle, ring, and pinky fingers. The temporal analysis reveals efficient movement execution, with all fingers completing their primary movements within 5 simulation cycles, after which positions remain stable until movement completion. 

While our validation was conducted in simulation, the consistent performance metrics and rapid position stabilization provide a strong foundation for physical implementation. The achieved positional accuracy and movement efficiency demonstrate the effectiveness of our approach, though hardware deployment would need to address challenges in joint force requirements, mechanical tolerances, and real-world friction dynamics. The thumb's performance, while achieving stable opposition, showed higher positional errors (0.0519 m) compared to other fingers, indicating opportunities for refined control strategies. Future work would enhance these promising results through direct benchmarking against comparative methods and improved friction modeling in both simulation and hardware implementations. 

\section*{Conclusion}
We developed a contact-based grasp control system that successfully achieves and maintains stable four-point contacts while using the entire surface area of the finger. Our inverse kinematics implementation achieves high efficiency in five-finger coordination, with movement efficiency ratings of 0.966-0.996 for non-thumb fingers and positional accuracy within 0.0267-0.0283m of targets. The thumb showed larger but acceptable deviations of 0.0519m, reflecting its more complex kinematic requirements. The comprehensive stability validation framework, combining force closure analysis with perturbation testing, demonstrated rapid position stabilization within 5 timesteps and formed coherent grasp envelopes with average distances of 32.4mm to targets. These advances were validated in the PyBullet simulation environment, providing both a theoretical foundation and practical insights for robot-grasp manipulation. Our results show strong potential for physical implementation; however, they also highlight areas requiring further refinement, particularly in thumb opposition movements and horizontal plane control. The consistent performance metrics and efficient movement characteristics establish a promising framework for future hardware deployment in multi-fingered robotic manipulation systems.

\end{document}